\documentclass{article}
 \usepackage[preprint]{neurips_2026}

\usepackage[utf8]{inputenc} 
\usepackage[T1]{fontenc}    
\usepackage{hyperref}       
\usepackage{url}            
\usepackage{booktabs}
\usepackage{xspace}
\usepackage{amsfonts}       
\usepackage{nicefrac}       
\usepackage{microtype}      
\usepackage{xcolor}         
\usepackage{graphicx}
\usepackage{natbib}
\usepackage{multicol}
\usepackage{multirow}
\usepackage{amsmath}
\usepackage[table]{xcolor}
\usepackage{wrapfig}
\usepackage{tikz}
\usepackage[most]{tcolorbox}
\usepackage{alltt}
\usepackage{enumitem}
\usepackage{algorithm}
\usepackage{amssymb}         
\usepackage{algpseudocode}   
\usepackage{amsmath}         
\usepackage{cleveref}        
\usepackage{pifont}
\usepackage{tablefootnote}
\usepackage{subcaption}

\newcommand{\cmark}{\ding{51}}
\newcommand{\xmark}{\ding{55}}

\tcbset{
  aibox/.style={
    width=\linewidth,
    top=0pt,
    colback=white,
    colframe=black,
    colbacktitle=black,
    enhanced,
    center,
    attach boxed title to top left={yshift=-0.1in,xshift=0.1in},
    boxed title style={boxrule=0pt,colframe=white,},
  }
}

\usepackage{xspace}
\newcommand{\ours}{CodeClinic\xspace}

\newtcolorbox{AIbox}[2][]{aibox,title=#2,#1}

\title{CodeClinic: Evaluating Automation of Coding Skills for Clinical Reasoning Agents}

\author{%
  \begin{tabular}{c}
    Timothy Ossowski\textsuperscript{1} \quad
    Xinchu Liu\textsuperscript{1} \quad
    Danyal Maqbool\textsuperscript{1} \quad
    Vaibhav Dhanuka\textsuperscript{1} \\[3pt]
    Sheng Zhang\textsuperscript{2} \quad
    Hoifung Poon\textsuperscript{2} \quad
    Majid Afshar\textsuperscript{1} \quad
    Tyler Bradshaw\textsuperscript{1} \quad
    Junjie Hu\textsuperscript{1} \\[5pt]
    {\mdseries\textsuperscript{1}University of Wisconsin--Madison \quad
    \textsuperscript{2}Microsoft Research}
  \end{tabular}
}

\begin{document}

\maketitle

\begin{abstract}
  Clinical reasoning agents based on large language models (LLMs) aim to automate tasks such as intensive care unit (ICU) monitoring and patient state tracking from electronic health records (EHRs). Existing systems typically rely on manually curated clinical tools or skills for concepts such as sepsis detection and organ failure assessment. However, maintaining these tool libraries requires substantial expert effort, while zero-shot querying or code generation often produces inefficient and unreliable reasoning chains, especially under institution-specific clinical policies. We introduce \ours{}, a benchmark built on MIMIC-IV for evaluating whether LLM agents can synthesize and compose reusable clinical skills instead of relying on fixed toolboxes. The benchmark contains two complementary tasks: longitudinal ICU surveillance and compositional information seeking. The longitudinal setting simulates monitoring patient trajectories with structured decisions every four hours across 25 findings and eight clinical families, while the compositional setting spans 63k instances across 259 tasks in nine domains and is stratified by compositional dependency depth to evaluate increasingly complex multi-step reasoning. We further propose an offline autoformalization pipeline that converts natural-language clinical guidelines into reusable and verified Python skill libraries through iterative LLM refinement. Compared with zero-shot code generation, the resulting libraries improve consistency while reducing per-query token usage by up to 40\%.\footnote{Code to reproduce data and evaluation available at \url{https://github.com/tossowski/CodeClinic}}.

\end{abstract}


\section{Introduction}

Large language models (LLMs) have demonstrated remarkable potential for clinical reasoning, achieving expert-level performance on medical licensing examinations~\cite{singhal2023large} and showing strong capabilities across a diverse range of medical tasks~\cite{bedi2026holistic, wu2025towards, wang2025capabilities}. Recent works have extended these capabilities to agentic settings, where LLMs interact with electronic health record (EHR) databases to answer clinically meaningful questions~\cite{liao2026agentehr, tang2024medagents}. These agents typically generate code or queries to extract structured information from EHR systems and reason over retrieved data to support clinical decisions.

A common approach is to equip agents with a user-designed toolbox of domain-specific functions for recurring clinical computations. While toolboxes improve reliability, they do not scale well: designing and maintaining them requires sustained expert effort and their coverage is inherently limited by the author's foresight. The alternative is a single-tool setup where the agent has only a generic interface, such as a single \texttt{query\_db} function, without hospital-specific helper functions. However, relying only on this minimal single-tool setup can hurt both accuracy and cost, as the agent must repeatedly issue complex queries and re-derive intermediate results.

Beyond the toolbox size tradeoff, many clinically meaningful questions are not well-defined without additional context. Clinical concepts such as sepsis screening, organ failure scoring, and vasopressor thresholds are not universal, with  precise definitions varying across institutions and evolving with clinical guidelines over time. An agent answering such questions without access to the relevant institutional policies and practice recommendations faces an underspecified problem, and any benchmark grading its output against a single fixed ground truth will produce misleading accuracy estimates.

Thus, our benchmark aims to evaluate a central question: \textit{Given limited verification data, how well can LLMs synthesize modular, composable libraries that operate over EHR data while dynamically accounting for institutional definitions and policies?} If successful, such models could remove the expert-curation bottleneck by generating code that adapts to location-specific clinical and health system policies.

We introduce \ours{}, a benchmark built on MIMIC-IV~\cite{johnson2020mimic} to evaluate this capability through two complementary tasks. The first is longitudinal Intensive Care Unit (ICU) surveillance, where an agent tracks patient state over time and issues structured monitoring decisions at 4-hour checkpoints across ICU stays, covering 25 findings across eight clinical families. The task simulates a doctor engaging in ICU patient monitoring, requiring persistent state tracking, handling time-sensitive variables, and recomputing composite conditions (e.g., sepsis) over a 13-step horizon without access to future information. The second is compositional information seeking, spanning 63 clinical concepts across nine domains and 63,000 queries, stratified by dependency depth to measure multi-step reasoning.

As a baseline, we propose an offline autoformalization pipeline that converts natural-language clinical guidelines into a reusable, verified Python function library via iterative LLM refinement. This library is constructed once using limited supervision and reused across all queries, ensuring consistency and reducing redundant computation.

Our contributions are as follows:
\begin{itemize}[leftmargin=20pt]
\item \textbf{\ours\ (Benchmark)}: a longitudinal agentic automation task for bedside ICU monitoring, paired with a compositional information seeking task which together surface the failure modes of LLM agents relying on static toolboxes and zeroshot code generation.
\item \textbf{Benchmark Analysis (Findings)}: A systematic characterization of where and why current LLMs fail on compositional and longitudinal clinical reasoning.
\item \textbf{Clinical Autoformalization (Baseline)}: An automated pipeline that transforms natural-language clinical guidelines into a verified Python function library using limited verification data, demonstrating a scalable alternative to user-designed toolboxes and serving as a strong baseline for future work on \ours{}.
\end{itemize}

\section{Related Work}
\begin{table}[!h]
\centering

\resizebox{\textwidth}{!}{%
\begin{tabular}{lccccc}
\toprule
\textbf{Benchmark} &
\textbf{Code Gen.} &
\textbf{Scale ($>$10K)} &
\textbf{Agentic} &
\textbf{Compositional} &
\textbf{Longitudinal} \\
\midrule
EHRSQL~\cite{lee2022ehrsql}            & \cmark & \cmark & \xmark & \xmark & \xmark \\
EHRSHOT~\cite{wornow2023ehrshot}      & \xmark & \cmark & \xmark & \xmark & \xmark \\
EHRAgent~\cite{shi2024ehragent}       & \cmark & \xmark & \cmark & \xmark & \xmark \\
AgentClinic~\cite{schmidgall2024agentclinic} & \xmark & \xmark & \cmark & \xmark & \xmark \\
MedAgentGym~\cite{xu2025medagentgym}  & \cmark & \cmark & \cmark & \xmark & \xmark \\
MedAgentBench~\cite{kim2025medagentbench} & \xmark & \xmark & \cmark & \xmark & \xmark \\
EHRStruct~\cite{zhang2025ehrstruct}   & \cmark & \xmark & \xmark & \xmark & \xmark \\
AgentEHR~\cite{liao2026agentehr}      & \xmark & \cmark & \cmark & \xmark & \xmark \\
\midrule
\textbf{\ours{} (Ours)} & \cmark & \cmark & \cmark & \cmark & \cmark \\
\bottomrule
\end{tabular}%
}
\caption{Comparison of EHR benchmarks relevant to clinical reasoning and agentic evaluation.
\textbf{Code Gen.}: requires generating executable code or SQL.
\textbf{Agentic}: supports iterative agent loop with execution feedback.
\textbf{Compositional}: explicitly tests multi-hop reasoning over dependent clinical concepts.
\textbf{Longitudinal}: evaluates reasoning over patient state as it evolves over time.}
\vspace{-0.5cm}
\label{tab:benchmark_comparison}
\end{table}
\paragraph{EHR Databases and Benchmarks}
Most prior EHR benchmarks build on the MIMIC family~\cite{johnson2020mimic, johnson2018mimic}. Text-to-SQL efforts (MIMICSQL~\cite{wang2020treqs}, EHRSQL~\cite{lee2022ehrsql}, EHR-SeqSQL~\cite{bae2024ehrseqsql}) and broader QA benchmarks (DrugEHRQA~\cite{bardhan-etal-2022-drugehrqa}, EHRXQA~\cite{bae2023ehrxqa}, EHRNoteQA~\cite{kim2024ehrnoteqa}) cover constrained query spaces over structured tables, multimodal records, or notes. EHRSHOT~\cite{wornow2023ehrshot} adds point-in-time foundation-model prediction, and code-generation benchmarks (EHRAgent~\cite{shi2024ehragent}, EHRStruct~\cite{zhang2025ehrstruct}) frame EHR reasoning as iterative code generation. All evaluate static, individually posed queries with expert-curated toolboxes, and none jointly stratify multi-hop concept dependencies and time-evolving clinical state. Table~\ref{tab:benchmark_comparison} summarizes the comparison; \ours{} is the only benchmark to combine compositional dependency-depth stratification across 63 clinical concepts with longitudinal reasoning over full ICU trajectories.

\paragraph{LLM-Based Medical Agents}
A growing body of work builds LLM agents that interact with EHR data or simulated clinical environments~\cite{liao2026agentehr, tang2024medagents, hager2024evaluation, bani2025language, xu2025medagentgym, huang2025biomni, kim2025medagentbench, schmidgall2024agentclinic}, with broader medical evaluation suites such as MedHELM~\cite{bedi2026holistic} measuring static task performance. Two limitations recur: agents rely on hand-crafted tool environments whose completeness bounds the evaluation, and re-executing full tool chains per query incurs significant token overhead. Our work is orthogonal: instead of evaluating an agent inside a fixed tool environment, we ask how well LLMs can \emph{generate} that environment.

\paragraph{Autoformalization and Code-Augmented Reasoning}
Autoformalization translates informal statements into executable form, an approach originally developed for mathematics by translating theorems into Lean or Isabelle~\cite{wu2022autoformalization}. We adapt the idea to clinical guidelines, replacing proof checkers with clinical verification and targeting a Python function library. Clinical definitions vary across institutions; although the natural-language guidelines we use draw on general consensus literature, the autoformalization loop grounds them in institution-specific practice by verifying generated code against labeled cases drawn from the target EHR system, fixing the ambiguity that makes zeroshot evaluation unreliable. Our pipeline runs a ReACT-style~\cite{yao2023react} code-as-reasoning loop~\cite{gao2023pal, chen2021codex} \emph{offline} to produce a stable, reusable library, separating costly library construction from inference and avoiding the repeated token expenditure of agents that re-derive the same computations at query time.

\begin{figure}[!t]
    \centering
    
    \includegraphics[width=\textwidth]{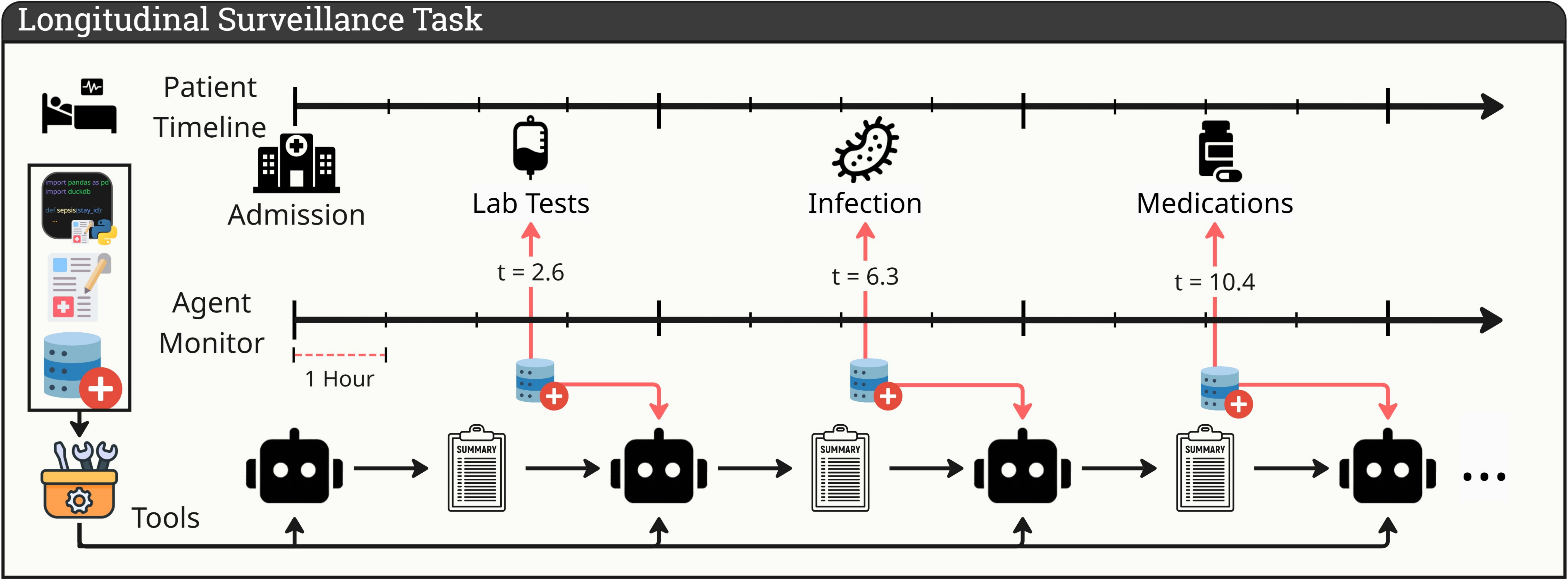}
    
    \caption{
    Illustration of the longitudinal ICU surveillance task in \ours{}. The agent is prompted every 4 hours throughout an ICU stay to assess the patient's current condition, with strict visibility constraints that limit it to data available up to the current checkpoint. At each checkpoint, the agent either reasons or writes code to gather evidence and outputs a surveillance decision. A rolling history of prior checkpoint decisions is carried forward across the stay via a summarization step.}
    \label{fig:longitudinal_task}
    
\end{figure}

\section{Benchmark}

\ours{} is built around two complementary tasks over MIMIC-IV~\cite{johnson2023mimic}: a \emph{longitudinal ICU surveillance} task (Figure \ref{fig:longitudinal_task}) that stresses agentic state tracking across rolling checkpoints, and a \emph{compositional information seeking} task (Figure \ref{fig:compositional_task}) that stresses multi-hop reasoning over the dependency graph of clinical concepts.

\subsection{Longitudinal ICU Surveillance}
\label{sec:bench-longitudinal}

The longitudinal task evaluates whether an agent can maintain clinically accurate state over time instead of answering a single isolated question. We draw a cohort of roughly $46$k ICU stays from MIMIC-IV by keeping only stays with at least $48$ hours of in-unit time, so that every trajectory is long enough to expose meaningful clinical evolution. Each stay is sliced into a fixed sequence of decision points spaced every four hours from admission through hour $48$, giving $13$ checkpoints per stay. Patients are assigned to splits at the subject level, so no patient appears in both the train and test set. From this cohort we release a held-out $2{,}000$ stay benchmark, sampled with a stratification which preserves the natural mix of ICU patients while preserving rarer high-acuity states.

\paragraph{Task Setup}
At each checkpoint, the agent emits a structured decision: a set of \emph{suspected conditions}, a set of \emph{alerts} that have crossed escalation thresholds, a single \emph{global action} (continue monitoring or escalate care), and a coarse \emph{priority} level. This compact interface is intentionally agent-friendly: it avoids committing to a fixed disease ontology while requiring the model to reason over the full multi-organ state at every step, rather than focusing on a single salient issue. The task also exhibits broad coverage. In the source cohort, over 90\% of stays trigger at least one surveillance family within the first day, and roughly two-thirds trigger three or more. The decision space spans both common and rare ICU conditions. The released $2{,}000$-stay benchmark preserves realistic co-occurrence across families while ensuring sufficient support for clinically important but less frequent conditions such as Acute Kidney Injury (AKI), septic shock, severe acidemia, and severe coagulopathy.

\begin{figure}
    \centering
    \includegraphics[width=\linewidth]{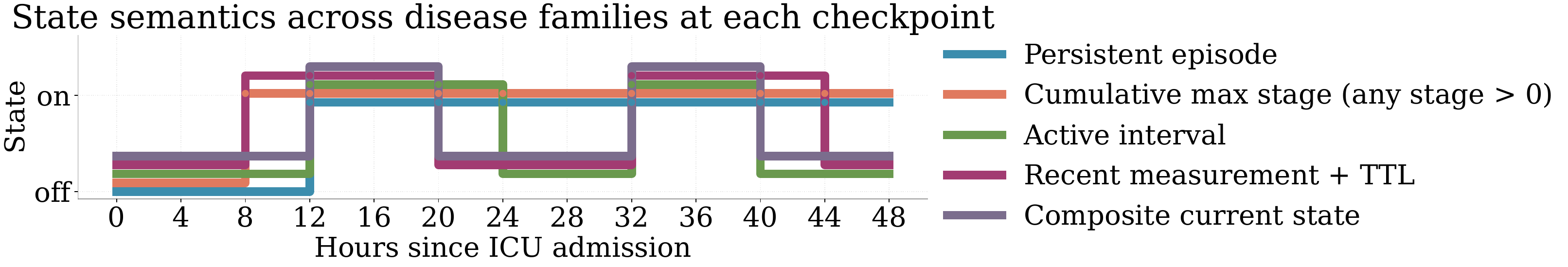}
    \vspace{-5mm}
    \caption{Example state dynamics across an ICU trajectory for different medical concepts. \ours{} challenges models to interpret these changes in state dynamics across the patient timeline. }
    \label{fig:surveillance_semantics}
\end{figure}

\paragraph{Label Generation}
Internally, we maintain a registry of $25$ canonical ICU surveillance findings, organized into eight clinical families that together cover the dominant ICU monitoring concerns: infection, sepsis, renal injury, respiratory support, hemodynamic support, neurologic deterioration, metabolic derangement, and coagulopathy. For each finding we define a deterministic builder over MIMIC-IV that consumes the patient's data up to the current checkpoint and decides whether the finding is currently active. The builder for each family encodes the appropriate clinical \emph{temporal semantics} (Figure~\ref{fig:surveillance_semantics}). Monotonic syndromes such as infection or sepsis remain active once they have been triggered within the trajectory. On the other hand, non-monotonic concepts such as ventilation or vasoactive support are active only while the intervention is in progress. A model therefore cannot solve the task by remembering a single binary label per disease. It must instead track which families persist once triggered, which families decay when evidence becomes stale, and which families are recomputed from multiple evolving components.



\paragraph{Agent Loop}
At each of the $13$ checkpoints the agent sees the current hour, a compact summary of its own decisions and observations from previous checkpoints, and access to a toolbox for grounding its predictions in the database. The benchmark supports several agent-facing configurations: an agent can be asked to decide each checkpoint independently or with rolling history. Tools can expose curated MIMIC-IV concept queries, the autoformalized function library produced by our pipeline (Appendix \ref{app:algorithm}), or only the raw EHR tables, allowing us to separately measure the contribution of longitudinal context and of higher-level clinical tooling. In all cases the agent must retrieve the evidence it deems relevant, commit to a structured decision for the current checkpoint, and write a short summary that will be visible at the next step.



\subsection{MIMIC-IV Compositional Information Seeking}

\begin{figure}[!t]
    \centering
    
    \includegraphics[width=\textwidth]{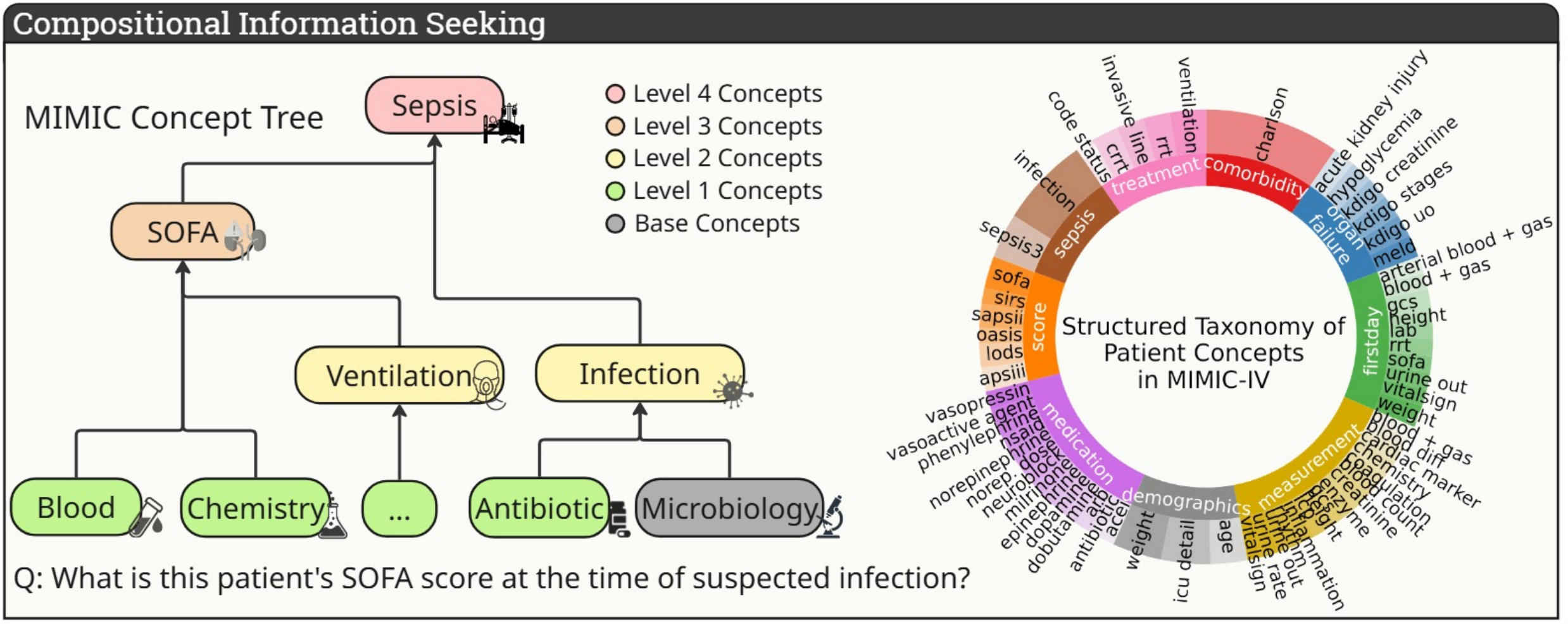}
    
    \caption{
\textbf{Left:} Example sepsis-related question with its corresponding dependency graph. Answering the query requires composing intermediate clinical concepts (e.g., Sequential Organ Failure Assessment (SOFA) score used in organ dysfunction definitions for sepsis, suspected infection) that must be recomputed over time.
\textbf{Right:} Hierarchical taxonomy of MIMIC clinical concepts, organized by supercategory and associated sub-concepts.
}
     \label{fig:compositional_task}
\end{figure}

We complement the longitudinal task with a clinical question answering benchmark grounded in the MIMIC-IV database. All questions are derived from tables in the MIMIC-IV \texttt{derived} schema, a curated collection of structured representations of the raw ICU data produced by community-validated SQL transformations applied to the original MIMIC-IV tables~\cite{johnson2021mimic}. The benchmark spans 63 clinical concepts covering nine physiological and pharmacological domains: comorbidity scoring, patient demographics, first-day assessments, laboratory measurements, medication administration, organ failure indices, severity scores, sepsis-related variables, and ICU treatments.

\paragraph{Concept Definition} Each concept corresponds to one derived table from MIMIC-IV (e.g., \texttt{sofa}) and is defined as a set of question variants paired with ground-truth extraction functions. Each concept specifies (i) a base SQL query that joins the derived table with \texttt{mimiciv\_icu.icustays} and \texttt{mimiciv\_hosp.admissions} to produce a per-stay feature matrix, and (ii) a set of question variants, each targeting a clinical attribute and question type. For each concept we also generate a natural-language guideline specification (defining clinical criteria, measurements, and time windows) by synthesizing PubMed literature with an LLM (details in Appendix \ref{app:guideline_gen}).

\paragraph{Dependency-Based Difficulty} A key property of MIMIC-IV derived tables is that many are computed from other derived tables, inducing a directed acyclic dependency graph. We assign each concept a difficulty level equal to its longest dependency-chain depth in this graph. Concepts with no upstream dependencies (e.g., \texttt{vitalsign}, \texttt{norepinephrine}) are \textit{Level 1}, representing raw chart or medication extractions. Depth-2 concepts (e.g., \texttt{ventilation}, \texttt{creatinine\_baseline}) are \textit{Level 2}, requiring one layer of aggregation. Concepts at depth $\geq$3 (e.g., \texttt{sofa}, \texttt{sepsis3}) are grouped as \textit{Level 3+}, reflecting composite clinical constructs that integrate multiple subsystems. This hierarchy serves as a proxy for reasoning difficulty, with higher levels requiring increasingly complex, multi-step clinical inference. We provide a visual representation of these dependencies in Figure \ref{fig:compositional_task}.

\paragraph{Clinical Knowledge Graph}
To determine dependency level difficulty, concepts are organized into a directed acyclic dependency graph $\mathcal{G} = (V, E)$, where an edge $(v_i, v_j)$ indicates that $v_j$ requires output from $v_i$. The graph is constructed by parsing SQL definitions in the MIMIC-Code repository and expanding with clinical conditions derived from ICD-10 diagnoses and UMLS Metathesaurus relations (see Appendix \ref{app:knowledge_graph} for full construction details). This structure drives both difficulty stratification and compositional function reuse in our baseline pipeline.


\textbf{Sampling and Splits} For each concept, up to 1000 ICU stays are sampled in total, distributed across its question variants. A 10\% subset (100 rows per variant) forms the training split used during autoformalization. We use a small training split to simulate a realistic scenario where labeled verification data is scarce and expensive. the remaining 90\% is the held-out test set. Applying this process results in 63,000 compositional queries across 42,311 unique ICU stays (6,300 train / 56,700 test). Ground-truth labels are extracted deterministically from derived tables.

\section{Baseline: Clinical Autoformalization}

\begin{figure}[!t]
    \centering
    
\includegraphics[width=\linewidth]{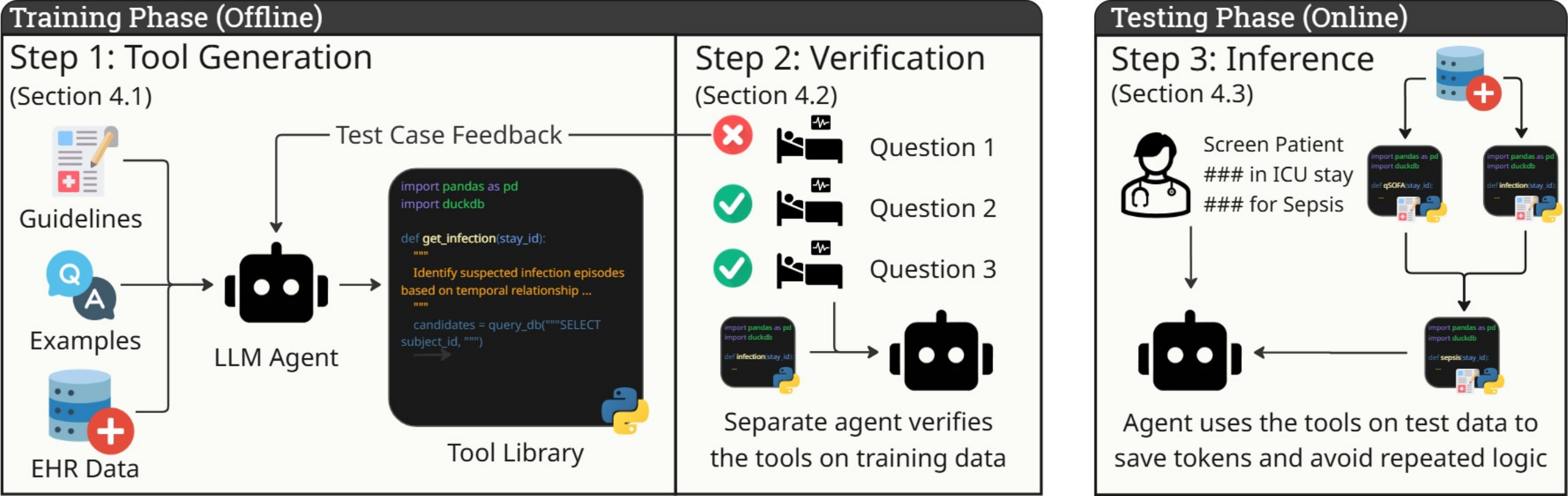}
    
    \caption{
Overview of the autoformalization pipeline. During the offline \textit{training phase}, an LLM agent uses medical guidelines, verification examples, and EHR data to construct a library of executable tools. A separate verification agent evaluates this library on the remaining training data, with errors fed back in an iterative refinement loop. During the online \textit{testing phase}, the learned tool library is provided to the inference agent, enabling more efficient computation and reduced token usage.
}
    \label{fig:pipeline}
\end{figure}

We propose a clinical autoformalization pipeline (Figure \ref{fig:pipeline}) that converts natural-language clinical guidelines into a verified, reusable Python function library using only a small train split as supervision. The pipeline takes as input a verification dataset $\mathcal{Q} = \{(q_k, y_k)\}_{k=1}^{K}$, an EHR database $\mathcal{C}$, and clinical guidelines $\mathcal{G}$, and produces a verified Python function $f$ that deterministically extracts a clinical concept for any patient. The full algorithm is given in Appendix \ref{app:algorithm}.

\subsection{Tool Generation}

The core of the pipeline is an iterative ReACT loop~\cite{yao2023react} in which an LLM agent explores the EHR database schema, consults clinical guidelines, and writes a Python function for a target concept. The agent operates within a stateful code interpreter exposing three tools: \texttt{query\_db(sql)} for database access, \texttt{search\_guidelines}/\texttt{get\_guideline} for retrieving clinical specifications from $\mathcal{G}$, and \texttt{search\_functions}/\texttt{load\_function} for discovering and importing previously verified functions. Guidelines are exposed as a searchable library rather than injected into the prompt, forcing the agent to bridge clinical knowledge and database schema through autonomous exploration.



\subsection{Verification}

Once the agent produces a candidate function $f_t$, it is evaluated against the 10\% train split of $\mathcal{Q}$. A separate agent calls $f_t$ in its own code interpreter and parses a verdict for each question to compare against ground truth. If accuracy exceeds $\theta = 0.90$, the function is accepted and saved to the shared library. Otherwise, per-question error traces are fed back to the autoformalization agent as structured feedback for the next iteration. The best-performing candidate across all iterations is retained, so the pipeline is robust to minor regressions between steps.

\subsection{Inference}

Once the library $\mathcal{L} = \{f_1, \dots, f_M\}$ is complete, it is deployed as a static resource over the held-out 90\% test and other clinical tasks. An LLM inference agent receives a question $q$ along with the available function signatures in $\mathcal{L}$ and access to the database $\mathcal{C}$, and iteratively selects and calls the functions it deems relevant, reasoning over their outputs to produce an answer:
\begin{equation}
    \hat{y} = \text{LLM}\bigl(q,\; \mathcal{L},\; \mathcal{C}\bigr)
\end{equation}
Because the functions in $\mathcal{L}$ have been verified on labeled cases, the inference agent's context is occupied by compact function signatures rather than raw schema exploration, substantially reducing token usage. The same function produces identical output for the same patient across runs, which is critical for reproducibility in clinical settings.

\begin{figure}[t]
\centering

\begin{minipage}[t]{0.4\linewidth}
\vspace{0pt}
\centering
\includegraphics[width=\linewidth]{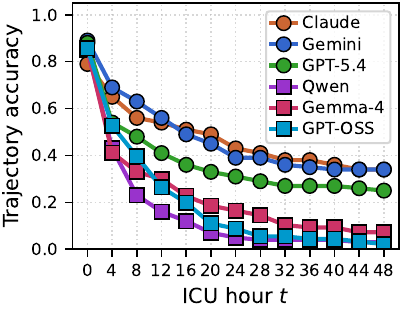}
\label{fig:your-figure}
\end{minipage}
\hfill
\begin{minipage}[t]{0.58\linewidth}
\vspace{0pt}
\centering
\small

\setlength{\tabcolsep}{4pt} 


\makebox[\linewidth]{%
\resizebox{\linewidth}{!}{%
\begin{tabular}{lrrrrr}
\toprule
Model & Acc. & Traj. Acc. & Prio. & Alerts F1 & Susp. F1 \\
\midrule
\multicolumn{6}{c}{\textbf{Frontier Models}\footnotemark[1]} \\
\midrule
Claude-Sonnet-4.6 & \textbf{76.2} & \textbf{34.0} & 56.8 & \textbf{47.2} & 38.3 \\
Gemini-3.1-Pro & 72.5 & 34.0 & \textbf{59.4} & 46.4 & \textbf{43.9} \\
GPT-5.4 & 66.5 & 25.0 & 51.7 & 32.1 & 25.6 \\
\midrule
\multicolumn{6}{c}{\textbf{Open-Source Models}} \\
\midrule
Qwen3.5-4B & 36.5 & 9.0 & 34.5 & \textbf{26.4} & 18.5 \\
Qwen3.5-9B & 33.0 & 7.7 & 31.4 & 23.8 & 20.4 \\
Qwen3.5-27B & 38.7 & 6.0 & 37.8 & 24.3 & 19.4 \\
Gemma-4-31B & 30.7 & 7.0 & 28.4 & 24.3 & 18.2 \\
GPT-OSS-120B & \textbf{64.5} & \textbf{12.0} & \textbf{47.9} & 24.7 & \textbf{21.6} \\
\bottomrule
\end{tabular}%
}%
}


\end{minipage}

\caption{\textbf{Left:} Trajectory accuracy on the rolling ICU surveillance task declines as monitoring windows lengthen. \textbf{Right:} Summary of longitudinal monitoring metrics (zeroshot method). }


\label{fig:rolling-comparison}
\end{figure}

\section{Results}

For our experiments we consider the following baselines. Each baseline is allowed up to 15 conversation turns of exploration (10 for longitudinal task) before being prompted to give a final answer. We provide an example trajectory in our environment in Appendix \ref{app:sample trajectory}:

\begin{enumerate}[label=\textbf{(\arabic*)},nosep,leftmargin=*]
    \item \textbf{Zeroshot}: The LLM receives only the raw MIMIC-IV schema and a \texttt{query\_db} function.
    
    \item \textbf{Clinical Autoformalization (Ours)}: The LLM is given the raw MIMIC-IV schema, a \texttt{query\_db} function, and the autoformalized function library $\mathcal{L}$ generated from the compositional information seeking train split. We use the same library created with \texttt{Qwen3.5-27B} across all runs.
\end{enumerate}

\paragraph{Evaluation Metrics: }
For the \textbf{longitudinal task}, each checkpoint is scored on global-action accuracy (\textbf{Acc}), priority level accuracy (\textbf{Prio}), and macro set-based $F_1$ over the predicted suspected-conditions and alerts sets (\textbf{Susp.\ F1}, \textbf{Alerts F1}). \textit{Trajectory accuracy} (\textbf{Traj.\ Acc}) is a stricter metric measuring the fraction of stays where the global action is correct at \emph{every} checkpoint, so a single error anywhere in the 13-step horizon counts as a full failure. For the \textbf{information seeking task}, answers are judged by exact match with 1\% tolerance, then macro-averaged across variants and concepts. \textbf{Tokens/Q} counts all tokens (prompt + generated) across all agent turns per question.

\subsection{Longitudinal Monitoring Results}



Figure ~\ref{fig:rolling-comparison} reports surveillance metrics and global-action accuracy as the trajectory unfolds. Note, we found that the GRPO finetuned model struggled to generalize to the new task, so we limit our analysis to zeroshot and autoformalization methods. We observed the following findings:

\textbf{Agents struggle to track patient state dynamics:} Even the strongest model (Claude-Sonnet-4.6) reaches only $34.0\%$ trajectory accuracy despite $76.2\%$ checkpoint accuracy. Our design forces the agent to keep monotonic syndromes (infection, sepsis) active once triggered while letting other states (oliguria, hyperlactatemia, severe acidemia) decay. Models that track only the latest checkpoint or only the running maximum are penalized by both regimes simultaneously.

\textbf{Disease tracking is harder than predicting actions: } Across all models, global-action accuracy ($\sim25$--$76\%$) is much higher than suspected-conditions $F_1$ ($\sim18$--$43\%$) and alerts $F_1$ ($\sim22$--$47\%$). These metrics probe whether the model can produce the \emph{full multi-organ picture} at each step rather than only the dominant problem, which is the clinically actionable signal in real ICU monitoring.
\footnotetext[1]{We ensured that no data was saved or used for training to comply with the MIMIC-IV data usage agreement. Frontier models also were run on a 21\% stratified subset of the data due to prohibitively expensive cost.}

\begin{figure}[!h]
    \centering
    \includegraphics[width=\linewidth]{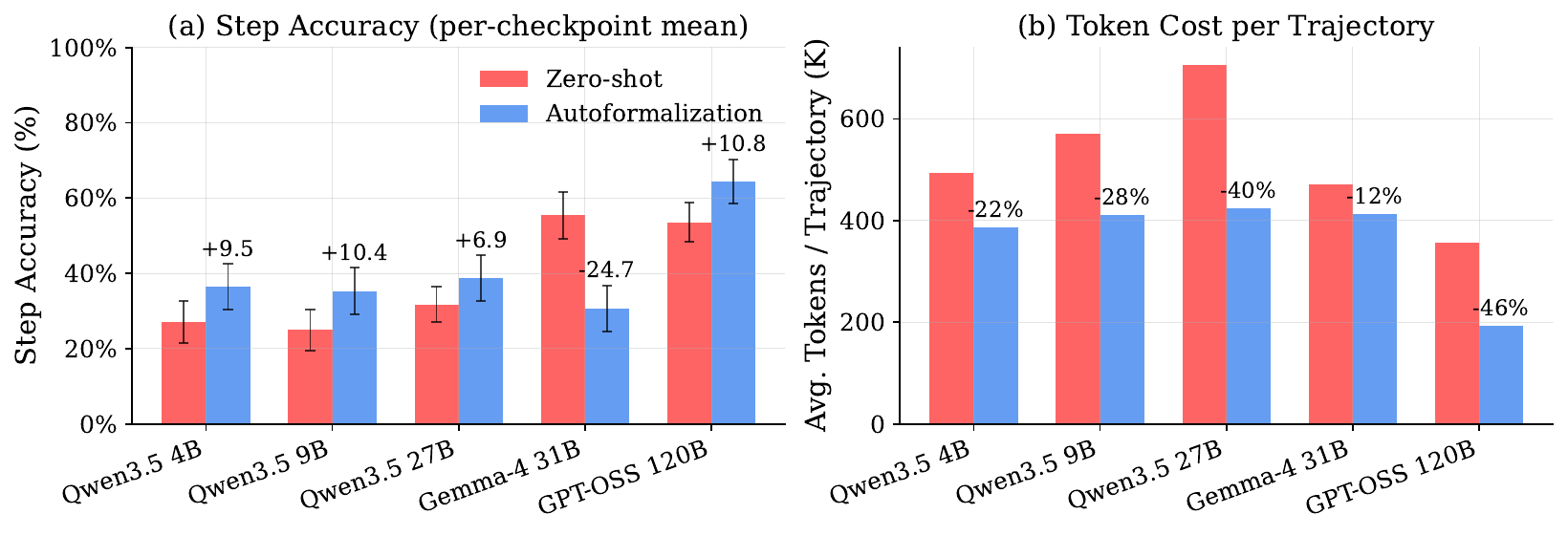}
    \vspace{-5mm}
    \caption{Step-level action prediction accuracy of the zeroshot versus autoformalization baseline comparison on the monitoring task. Error bars: 95\% Wilson score interval ($z=1/96$).}
    \label{fig:autoform_longitudinal}
\end{figure}

\textbf{Autoformalized functions improve longitudinal consistency.} As shown in Figure~\ref{fig:autoform_longitudinal}, the majority of models equipped with the autoformalized library achieve higher step accuracy than their zero-shot counterparts, with gains in token efficiency across all models. Because the pre-verified functions encode the correct temporal semantics of each clinical concept, the agent no longer needs to re-derive these rules from scratch at each checkpoint, reducing the per-step error rate.

\subsection{Compositional Information Seeking Analysis}
\label{sec:task_difficulty}

To characterize the difficulty of \ours{}'s compositional information seeking component, we evaluate a diverse set of models in a zero-shot code-generation setting. Each model is given the raw MIMIC-IV schema, a \texttt{query\_db} function, and a clinical question, and must iteratively generate executable code to retrieve and analyze information from the database. The model can then reason over execution outputs to produce an answer or continue refining its code, for up to 15 iterations. We select models spanning multiple families and scales to ensure broad coverage and to control for potential schema memorization from pretraining data, since MIMIC-IV is a widely studied dataset.

\begin{table}[!h]
\centering

\begin{tabular}{lcccccc}
\toprule
\multirow{2}{*}{Model} & \multicolumn{4}{c}{Accuracy by Difficulty Level} & \multirow{2}{*}{Overall} & \multirow{2}{*}{Tokens/Q} \\
\cmidrule(lr){2-5}
& Level 1 & Level 2 & Level 3+ & $\Delta$ (L1$\to$L3+) & & \\

\midrule
\multicolumn{7}{c}{\textbf{Frontier Models}\footnotemark[1]} \\
\midrule

Claude-Sonnet-4.6  & \textbf{58.4} & 51.0 & \textbf{41.1} & \textbf{-17.3} & \textbf{53.1} & 80,308 \\
Gemini-Pro-3.1  & 58.0 & \textbf{54.1} & 37.2 & -20.8 & 53.0 & \textbf{15,897}\\
GPT-5.4  & 55.1 & 47.8 & 36.7 & -18.4 & 49.6 & 16,778 \\

\midrule
\multicolumn{7}{c}{\textbf{Open-Source Models}} \\
\midrule

Gemma4-31B        & \textbf{46.9} & \textbf{44.2} & \textbf{32.2} & -14.7 & \textbf{43.4} & 39,243 \\
gpt-oss-120B      & 39.4 & 38.6 & 23.2 & -16.2 & 36.1 & \textbf{19,576} \\
Qwen3.5-27B       & 36.5 & 38.6 & 21.6 & -14.9 & 34.2 & 34,414 \\
Qwen3.5-9B        & 22.8 & 22.9 & 8.7 & \textbf{-14.2} & 20.2 & 41,378 \\
Qwen3.5-4B        & 25.4 & 20.9 & 8.6 & -16.8 & 21.0 & 49,445 \\

\bottomrule
\end{tabular}

\footnotetext{We ensured that no data was saved or used for training by these model APIs.}
\vspace{2mm}
\caption{Zeroshot accuracy (\%) of LLMs on the \ours{} compositional information seeking task by difficulty level. All models are given only the raw MIMIC-IV schema and a \texttt{query\_db} tool.}
\vspace{-3mm}
\label{tab:llm_difficulty}
\end{table}

Table~\ref{tab:llm_difficulty} reports accuracy across 63 concepts by difficulty level. Several trends emerge. First, even the strongest frontier models achieve only moderate accuracy, confirming that clinical information seeking over real EHR databases remains challenging. The best-performing model (Claude-Sonnet-4.6) reaches 53.1\% overall accuracy, well below expert-level performance, indicating substantial headroom for future systems. Second, performance consistently declines from Level~1 to Level~3+ across all model families, with the drop (column $\Delta$) reflecting the compounding effect of multi-hop dependencies. For example, answering a Sepsis-3 question requires correctly computing SOFA sub-scores, infection criteria, and antibiotic timing, each of which can fail independently.

\begin{table}[!h]
\centering
\resizebox{\linewidth}{!}{
\begin{tabular}{lcccccc}
\toprule
\multirow{2}{*}{Method} & \multicolumn{3}{c}{Accuracy by Level} & \multirow{2}{*}{Overall Acc.} & \multirow{2}{*}{Tokens/Q} & \multirow{2}{*}{Time/Q (s)} \\
\cmidrule(lr){2-4}
& Level 1 & Level 2 & Level 3+ & & & \\
\midrule
 Zeroshot         & 44.9 & 32.8 & 26.3 & 38.1 & 30479 & 3.1 \\
GRPO Finetuning                    & \textbf{57.1} & 36.5 & 27.9 & 40.5 & \textbf{22376} & 2.7 \\
Autoformalization (Ours)                 & 51.2 & \textbf{41.5} & \textbf{34.1} & \textbf{45.3} & 23569 & \textbf{2.5} \\
\bottomrule
\end{tabular}
}
\vspace{1mm}
\caption{Comparison of approaches across all 63 concepts on the held-out compositional information seeking test set. Accuracy is macro-averaged across concepts. Token usage and time are reported per question at inference.}
\label{tab:baselines}
\end{table}

\paragraph{Comparison with Finetuning}
\label{sec:baselines}

Table~\ref{tab:baselines} compares a traditional RL finetuning approach against our autoformalization pipeline, alongside a zeroshot baseline, to isolate the effect of learning reusable code libraries versus adapting the model weights directly. All approaches use \texttt{Qwen3-4B} as the backbone LLM. For GRPO finetuning, we finetune \texttt{Qwen3.5-4B} \cite{yang2025qwen3} using standard GRPO \cite{shao2024deepseekmath} with 50 training examples per concept and rewards based on exact match between the predicted and ground truth answer. Finetuning improves Level 1 accuracy, likely by teaching shallow database search patterns, but performance drops substantially on Level 2+ tasks requiring deeper compositional reasoning. In contrast, our autoformalization pipeline achieves performance competitive with finetuning without parameter updates, demonstrating that the offline verification loop can synthesize useful reusable functions. Autoformalization also reduces per-query token usage by approximately 23\% compared to zeroshot inference, since the agent invokes compact verified functions instead of repeatedly rediscovering schema logic. Full training details are provided in Appendix~\ref{app:train_details}.

\begin{figure}[!h]
    \centering
    \includegraphics[width=\linewidth]{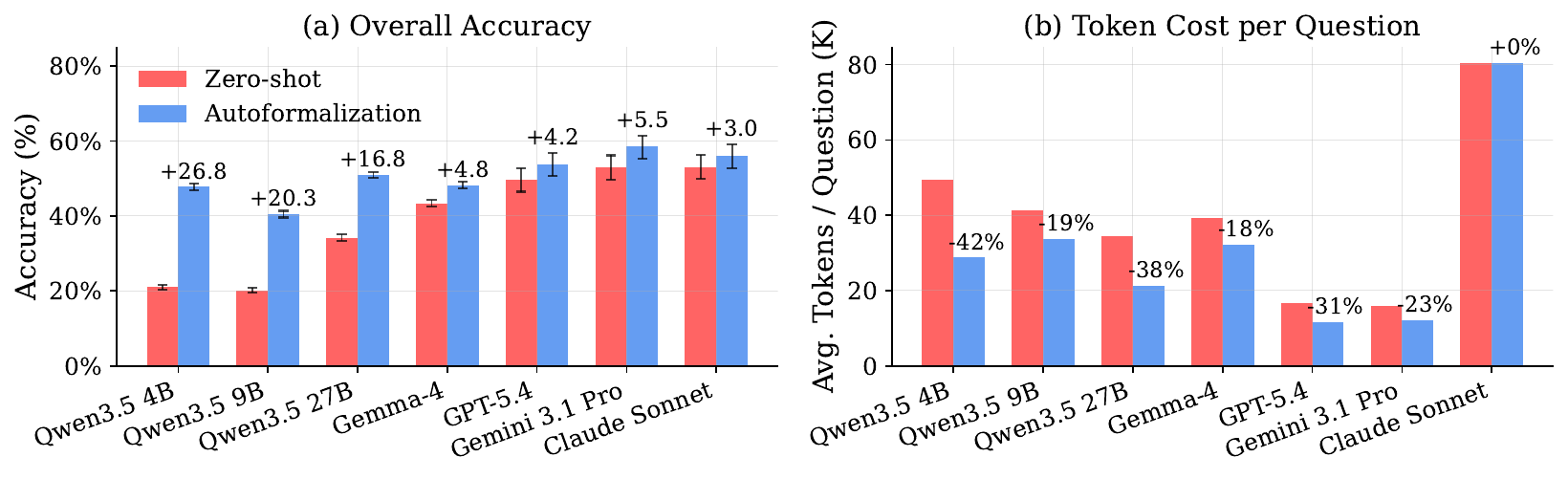}
    \vspace{-5mm}
    \caption{Performance comparison between zeroshot evaluation and our autoformalization baseline on the compositional information seeking task. Autoformalization shows universal performance improvement while improving or matching token efficiency across all models. Error bars: 95\% bootstrapped interval across 2000 resamples.  }
    \label{fig:comparison_overall}
\end{figure}

\textbf{Autoformalization Details } As observed in the longitudinal task, Figure~\ref{fig:comparison_overall} illustrates that accuracy gains from autoformalization in information seeking are consistent across model families, with the relative benefit most pronounced for smaller models which frequently fail to chain dependent clinical computations correctly. Token efficiency similarly improves or matches zeroshot across models, confirming that the offline verification cost is amortized over many queries and does not inflate per-question inference cost. The cross-model consistency of these gains suggests that the benefit derives from the verified functions themselves and motivates autoformalization as a general-purpose approach for grounding LLM agents in clinical code libraries.

\section*{Conclusion}

We presented \ours{}, an agentic benchmark consisting of 65k coding instances spanning 260 tasks over 63 clinical concepts, designed to evaluate whether LLMs can generate modular, composable libraries for EHR data processing. As a strong baseline, we proposed clinical autoformalization, which converts natural-language clinical guidelines into verified Python libraries via a QA-driven refinement loop, removing the need for manual tool engineering. The resulting libraries also generalize to the longitudinal task and reduce per-query token usage compared to zeroshot approaches. We see opportunity in extending \ours{} to multimodal data sources and in using the verified libraries for deploying smaller clinical agents for realistic ICU monitoring workflows.

\bibliographystyle{unsrtnat}
\bibliography{refs}

\appendix
\clearpage

\section{Limitations}

Several limitations bound the current work. First, the benchmark is grounded in MIMIC-IV, a dataset from a single US academic medical center, and the autoformalized functions inherit the clinical conventions of that institution. Generalization to EHR systems with different schema designs, coding practices, or patient populations is not guaranteed without re-running the autoformalization loop. Second, the verification set used during autoformalization is small (approximately 100 examples per concept), which may allow rare edge cases to escape detection, although we intentionally design this split to mirror the scarcity and expensive nature of real medical data. Third, the natural-language guidelines used as input are generated by an LLM from PubMed abstracts. Although we manually inspected the summaries for quality, this adds a layer of potential drift from ground-truth clinical documents.

\section{Broader Impacts}

This work has potential benefits for clinical informatics by reducing the expert effort required to deploy LLM-based EHR analysis tools and by promoting reproducible, verifiable code over zeroshot generation. A library of verified clinical functions is more trustworthy than a sequence of one-off model outputs, which is relevant for safety-critical applications. At the same time, automated extraction of clinical concepts from EHR data carries risks if deployed without appropriate clinical validation: errors in autoformalized functions could propagate silently to clinical decision support systems. Additionally, MIMIC-IV is demographically skewed toward patients treated at Beth Israel Deaconess Medical Center in Boston, and function libraries trained on this distribution may perform less reliably for underrepresented patient groups. We recommend that any downstream deployment of autoformalized libraries be subject to institution-specific validation before clinical use.

\section{Benchmark Statistics}
\label{app:benchmark_stats}
\begin{table}[h]
\centering
\small
\setlength{\tabcolsep}{4pt}
\caption{Compositional benchmark statistics.}
\label{tab:comp-stats}

\begin{tabular}{lcccc}
\toprule
            & Level 1 & Level 2 & Level 3+ & Total \\
\midrule
Concepts    & 34      & 17      & 12       & 63    \\
\bottomrule
\end{tabular}

\vspace{6pt}

\begin{tabular}{lrp{0.55\linewidth}}
\toprule
Question type & Count & Example question \\
\midrule
Comparative & 16{,}800 & By how many ng/mL does this patient's maximum troponin T level exceed the myocardial injury threshold of 0.1 ng/mL? \\
Aggregation & 15{,}600 & What was this patient's maximum milrinone infusion rate (mcg/kg/min) during their ICU stay? \\
Temporal    & 14{,}400 & How many hours after ICU admission was this patient first started on CRRT? \\
Direct      & 6{,}000  & What is this patient's GCS motor component score on their first ICU day? \\
Count       & 4{,}200  & How many distinct episodes of invasive mechanical ventilation did this patient have during their ICU stay? \\
Derived     & 3{,}400  & What is the interval in hours between this patient's first antibiotic and first culture? \\
Ratio       & 1{,}800  & What is this patient's BUN-to-creatinine ratio on their first ICU day? \\
Arithmetic  & 800      & In which decade of life is this patient? \\
\bottomrule
\end{tabular}
\end{table}

\begin{table}[t]
\centering
\small
\setlength{\tabcolsep}{4pt}
\label{tab:long-stats}

\begin{tabular}{lccccc}
\toprule
       & Trajectories & Checkpoints & Interval & Horizon & Median ICU LOS \\
\midrule
Cohort & 2{,}000 & 26{,}000 & 4\,h & 48\,h (13 ckpts) & 93.6\,h \\
\bottomrule
\end{tabular}

\vspace{6pt}

\begin{tabular}{lcc}
\toprule
        & Continue monitoring & Escalate \\
\midrule
Action  & 5{,}886 & 20{,}114 \\
\bottomrule
\end{tabular}
\hfill
\begin{tabular}{lccc}
\toprule
         & Low & Medium & High \\
\midrule
Priority & 5{,}408 & 10{,}585 & 10{,}007 \\
\bottomrule
\end{tabular}

\vspace{6pt}

\end{table}

\begin{table}[t]
\centering
\small
\caption{Longitudinal ICU surveillance benchmark statistics. Counts are over the 26{,}000 checkpoints in the released 2{,}000-stay package unless otherwise noted.}
\label{tab:long-stats}
\begin{tabular}{@{}lr@{\hskip 1.5em}lr@{\hskip 1.5em}lr@{}}
\toprule
\multicolumn{2}{c}{\emph{Cohort}} &
\multicolumn{2}{c}{\emph{Label distribution}} &
\multicolumn{2}{c}{\emph{Active clinical families}} \\
\cmidrule(lr){1-2}\cmidrule(lr){3-4}\cmidrule(lr){5-6}
ICU trajectories     & 2{,}000          & \multicolumn{2}{l}{\textit{Global action}} & Infection    & 19{,}984 \\
Checkpoints          & 26{,}000         & \quad Continue monitoring & 5{,}886        & Renal        & 16{,}568 \\
Checkpoint interval  & 4\,h             & \quad Escalate            & 20{,}114       & Sepsis       & 14{,}585 \\
Horizon per stay     & 48\,h (13 ckpt.) & \multicolumn{2}{l}{\textit{Priority}}      & Respiratory  & 9{,}874  \\
Median ICU LOS       & 93.6\,h          & \quad Low                 & 5{,}408        & Hemodynamic  & 5{,}765  \\
                     &                  & \quad Medium              & 10{,}585       & Coagulation  & 5{,}505  \\
                     &                  & \quad High                & 10{,}007       & Metabolic    & 4{,}696  \\
                     &                  &                           &                & Neurologic   & 1{,}389  \\
\bottomrule
\end{tabular}
\end{table}

\clearpage
\section{Guideline Generation from Medical Literature}
\label{app:guideline_gen}

For each clinical concept in the dependency graph, we generate a structured natural-language specification by querying PubMed. Given a concept name (e.g., ``acute kidney injury''), we search PubMed for relevant clinical practice guideline articles and systematic reviews, retrieve 3--5 top-ranked results, and extract their abstracts and available full text. An LLM then synthesizes the retrieved literature into a structured specification containing:
\begin{enumerate}[label=(\roman*),nosep]
    \item the clinical definition and diagnostic criteria with explicit measurement thresholds (e.g., serum creatinine increase $\geq 0.3$~mg/dL within 48 hours per KDIGO guidelines),
    \item the required EHR measurements mapped to MIMIC-IV table names and item identifiers,
    \item severity staging criteria where applicable, and
    \item temporal windows for event detection.
\end{enumerate}
This pipeline produces guideline specifications for $N$ concepts spanning the measurement, scoring, and condition layers of the dependency graph.

\section{Clinical Knowledge Graph Construction}
\label{app:knowledge_graph}

We construct a directed acyclic dependency graph $\mathcal{G} = (V, E)$ where an edge $(v_i, v_j) \in E$ indicates that concept $v_j$ requires the output of concept $v_i$. The graph is built in three phases.

\paragraph{Phase 1: Infrastructure nodes.} We parse SQL definitions from the MIMIC-Code repository, extracting references to upstream derived tables, raw MIMIC-IV tables, and \texttt{itemid} values via regex matching to produce infrastructure nodes (measurements, medications, severity scores, treatments) with exact computational dependencies.

\paragraph{Phase 2: Clinical conditions.} We expand the graph by mining frequent ICD-10 diagnosis codes among MIMIC-IV ICU patients, filtering for conditions detectable from structured EHR data, and generating guideline specifications using the pipeline in Appendix~\ref{app:guideline_gen}. Dependencies for each condition are inferred by keyword-matching the specification text against infrastructure node descriptions and entity labels.

\paragraph{Phase 3: UMLS expansion.} We map existing nodes to UMLS Concept Unique Identifiers and traverse one-hop Metathesaurus relations, filtering candidates by semantic type and ranking by centrality (the number of seed nodes connecting to each candidate).

The resulting graph contains multiple quality tiers---from infrastructure nodes with exact SQL-parsed dependencies, through conditions with guideline-validated dependencies, to automatically discovered conditions with heuristic dependency assignments---enabling both precise retrieval for core clinical concepts and scalability evaluation across a broad condition library.

\clearpage


\section{Autoformalization Algorithm}
\label{app:algorithm}

\begin{algorithm}[h]
\caption{Autoformalize}
\label{alg:autoformalize}
\begin{algorithmic}[1]
\Require verification dataset $\mathcal{Q}$, database $\mathcal{C}$, guidelines $\mathcal{G}$, LLM $\mathcal{M}$, threshold $\theta$, max iterations $T$
\State $\mathcal{S} \leftarrow \textsc{CodeSession}(\mathcal{C}, \mathcal{G})$; \;
       $f^* \leftarrow \emptyset$; \; $\alpha^* \leftarrow 0$; \; $\text{mem} \leftarrow []$
\State $\mathit{msgs} \leftarrow [\textsc{SysPrompt}, \textsc{UserPrompt}(\mathcal{Q})]$
\For{$t = 1, \dots, T$}
    \If{context near limit}
        \State $\text{mem} \leftarrow \textsc{Compress}(\text{mem} \cup \textsc{Summarize}(\mathit{msgs}))$
        \State $\mathit{msgs} \leftarrow [\textsc{SysPrompt}, \textsc{MemResume}(\text{mem})]$
    \EndIf
    \State $r \leftarrow \mathcal{M}(\mathit{msgs})$; \; $\mathit{msgs} \mathrel{+}= [r]$
    \State $\mathcal{B} \leftarrow \textsc{ExtractCodeBlocks}(r)$; \; \textbf{if} $\mathcal{B}=\emptyset$ \textbf{continue}
    \State $\mathit{msgs} \mathrel{+}= [\textsc{Execute}(\mathcal{B}, \mathcal{S})]$
    \If{$\texttt{FINAL\_FUNCTION} \notin \mathcal{S}.\text{namespace}$}
        \State \textbf{continue}
    \EndIf
    \State $f \leftarrow \mathcal{S}.\text{namespace}[\texttt{FINAL\_FUNCTION}]$
    \State $(\alpha, \mathcal{E}) \leftarrow \textsc{Eval}(f, \mathcal{Q}, \mathcal{C}, \mathcal{M})$
    \If{$\alpha > \alpha^*$} \; $f^* \leftarrow f$; \; $\alpha^* \leftarrow \alpha$ \EndIf
    \If{$\alpha \geq \theta$} \textbf{break} \EndIf
    \State $\mathit{msgs} \mathrel{+}= [\textsc{Feedback}(\alpha, \theta, \mathcal{E})]$
\EndFor
\State \Return $f^*$
\end{algorithmic}
\end{algorithm}

\paragraph{Context Compression.}
Because the ReACT loop may run for up to $T_{\max}$ iterations (default 100) with substantial code and query outputs at each step, the conversation history can exceed the LLM's context window. We address this with a rolling memory mechanism: when the estimated token count exceeds 25{,}000 tokens, the accumulated conversation is summarized by a separate LLM call into a compact memory document retaining key schema facts, the latest function source code, and a record of test results and error fixes. The full conversation history is then replaced by a two-message prompt (system instructions $+$ memory resume). This compression is lossy but preserves the information most critical for continued refinement.

\clearpage


\section{Prompts}
\label{app:prompts}

We document four prompt templates used throughout the paper: the autoformalization agent prompts (system and user, used to construct the Python concept library) and the evaluation prompts (system and user, used to score the generated functions on held-out questions). We then document the two prompt templates used by the longitudinal ICU surveillance agent at each 4-hour checkpoint.

\subsection{Autoformalization Agent}
\begin{figure*}[h]
    \begin{AIbox}{Autoformalization System Prompt}

\begin{quote}
\ttfamily
\tiny
\vspace{2mm}

You are an expert clinical informaticist and Python programmer.

You can execute Python code by placing it inside <code> ... </code> XML tags.
The code runs in a persistent session with these pre-loaded:
  - `query\_db(sql)` — runs a SQL query on the DuckDB database, returns a
    pandas DataFrame
  - `pandas` (pd), `numpy` (np), `duckdb`, `datetime` are pre-imported
  - `DB\_PATH` — path to the DuckDB database

IMPORTANT DATABASE NOTES:
  
  \{Database Specific Notes\}

YOUR TASK:
Write a Python function that extracts critical information for the clinical concept described
in the guideline you are given.  The function should use `query\_db()` to
look up patient data from the database and return useful clinical information.

Steps:

1. **Explore the database schema** — list tables, inspect columns, look up
   dictionary/lookup tables (d\_labitems, d\_items, d\_icd\_diagnoses, etc.).
   Run sample queries to understand data formats.

2. **Write a Python function** — implement the guideline's logic.  The
   function should accept patient identifiers (e.g. subject\_id, hadm\_id,
   stay\_id) and return information that captures the clinical concept.
   The output format is flexible — return whatever best represents the
   concept (e.g., a value, dict, DataFrame, boolean, list, etc.).

3. **Test thoroughly** — call your function with several sample patients
   and verify the output looks reasonable.  Debug and fix any issues.
   Do NOT assign FINAL\_FUNCTION until you are confident it works.

4. **Save it** — once you are satisfied, write a **single, self-contained
   code block** that includes ALL imports, helper functions, and the main
   function definition, then assigns FINAL\_FUNCTION at the end.  This
   block must be runnable on its own — do NOT include test calls, print
   statements, or assertions in it.  
   
   Example:

       import pandas as pd

       def helper(x):
           ...

       def compute\_concept(subject\_id, hadm\_id):
           ...

           return value

       FINAL\_FUNCTION = compute\_concept

5. After you receive evaluation feedback, **revise** your function,
   test the revision, then submit a new self-contained code block
   assigning FINAL\_FUNCTION again.

IMPORTANT RULES:

- Your function MUST use `query\_db()` to access the database.

- Do NOT call `duckdb.connect()` directly — it will cause connection errors.

- Your function should accept patient identifiers and return information
  that usefully captures the clinical concept. Any output format is fine.
  
- The function must work for any valid patient, not just the test cases.

- Do NOT guess column names or item IDs — always look them up first.

- The code block that assigns FINAL\_FUNCTION must be entirely
  self-contained (all needed imports, helpers, and the function itself)
  and must NOT contain any test calls, print statements, or assertions.

Be methodical.  Explore first, code second.  
You may include multiple <code> blocks in a single response.

\end{quote}

\end{AIbox}
    \caption{The system prompt provided to the LLM agent during the autoformalization react loop. The prompt can be customized to include database specific information, such as important column names, primary key, etc.}
    \label{prompts:system_prompt}
    
\end{figure*}
\begin{figure*}[h]
    \begin{AIbox}{Autoformalization Task Prompt}

\begin{quote}
\ttfamily
\tiny
\vspace{2mm}

YOUR TASK --- build a function for the clinical concept `\{concept\_name\}':

You need to write a Python function that can help answer clinical
questions about `\{concept\_name\}' for individual patients.

Here are example questions your function will need to support:
\{sample\_questions\}

APPROACH:
1. **Search for guidelines** --- call `search\_guidelines()' to see what clinical
   guidelines are available, then use relevant keywords (e.g.
   `search\_guidelines("\{concept\_name\}")') to find the most relevant ones.
   Read each relevant guideline with `get\_guideline(name)' --- these contain
   clinical definitions, scoring criteria, and implementation details that
   are essential for getting the logic right.
2. **Check for reusable functions** --- call `search\_functions()' to see if any
   previously implemented functions could serve as helpers. Use
   `get\_function\_info(name)' to inspect their signatures and docstrings,
   and `load\_function(name)' to import ones you want to reuse.
3. **Explore the database schema** --- list tables, inspect columns, look up
   dictionary/lookup tables (d\_labitems, d\_items, d\_icd\_diagnoses, etc.).
   Run sample queries to understand data formats.
4. **Write a Python function** --- implement the clinical logic based on what
   you learned from the guidelines and the database schema.  The function
   should accept patient identifiers (e.g. subject\_id, hadm\_id, stay\_id)
   and return information that captures the clinical concept.
   The output format is flexible --- return whatever best represents the
   concept (e.g., a value, dict, DataFrame, boolean, list, etc.).
5. **Test thoroughly** --- call your function with several sample patients
   and verify the output looks reasonable.  Debug and fix any issues.
   Do NOT assign FINAL\_FUNCTION until you are confident it works.
6. **Save it** --- once satisfied, write a single self-contained code block
   with ALL imports, helpers, function def, and FINAL\_FUNCTION assignment.
   No test calls or print statements in that final block.

FINAL FUNCTION REQUIREMENTS:
  - It should return useful clinical information that captures the concept.
  - The input should involve some patient identifier (e.g. stay\_id)
  - The output can be any format (scalar, dict, DataFrame, etc.) --- choose
    whatever best represents the clinical concept.
  - Use `query\_db(sql)' or `query\_fhir(resource\_type, params)' to look up data.
  - Have a detailed docstring which explains the expected input and output.

Start by searching for guidelines relevant to `\{concept\_name\}'.

\end{quote}

\end{AIbox}
    \caption{The task prompt provided to the LLM agent to initiate autoformalization of a clinical concept. Template variables \{concept\_name\} and \{sample\_questions\} are filled at runtime.}
    \label{prompts:react_user_prompt}

\end{figure*}

\subsection{Compositional Information Seeking Evaluation}
\begin{figure*}[h]
    \begin{AIbox}{Compositional Information Seeking Evaluation System Prompt}

\begin{quote}
\ttfamily
\tiny
\vspace{2mm}

You are a clinical expert evaluating patient data for clinical concepts.

Execute Python code inside <code> ... </code> tags. The session pre-loads:
  - `query\_db(sql)' --- SQL query $\rightarrow$ pandas DataFrame
  - The concept function described in your task
  - `pandas' (pd), `numpy' (np), `datetime' pre-imported

Database: DuckDB with schemas `mimiciv\_hosp', `mimiciv\_icu', `mimiciv\_derived'.
Always use fully-qualified table names (e.g. `mimiciv\_hosp.admissions').
Never call `duckdb.connect()' --- use `query\_db(sql)'.
IDs: subject\_id (patient) $\rightarrow$ hadm\_id (admission) $\rightarrow$ stay\_id (ICU stay).

\end{quote}

\end{AIbox}
    \caption{System prompt for the compositional information seeking evaluation. The agent uses the autoformalized concept function loaded into the session to answer clinical questions about individual patients.}
    \label{prompts:qa_system_prompt}

\end{figure*}

\begin{figure*}[h]
    \begin{AIbox}{Compositional Information Seeking User Prompt}

\begin{quote}
\ttfamily
\tiny
\vspace{2mm}

Session tools:
- `query\_db(sql)' --- duckdb SQL query $\rightarrow$ pandas DataFrame
- `search\_functions(keyword="")' --- list available concept functions by name
- `load\_function(name)' --- load a concept function into the session
- `get\_function\_info(name)' --- view a function's signature and docstring
- `search\_guidelines(keyword="")' / `get\_guideline(name)' --- clinical guidelines

Concept functions relevant to this task are available --- use `search\_functions()' to discover them and `load\_function(name)' to load what you need. You should always check first to see if there are any useful functions for you to use.
\{func\_info\}
Patient: subject\_id=\{subject\_id\}, hadm\_id=\{hadm\_id\}, stay\_id=\{stay\_id\}

\{query\}

Use the available functions and/or `query\_db' to answer, then reply with: \{verdict\_format\}
For numerical questions, include as much decimal precision as possible.

\end{quote}

\end{AIbox}
    \caption{User prompt for the evaluation agent. Template variables are filled at runtime: \{func\_info\} provides signatures of relevant concept functions, \{subject\_id\}/\{hadm\_id\}/\{stay\_id\} identify the target patient, \{query\} is the clinical task, and \{verdict\_format\} specifies the answer format.}
    \label{prompts:qa_user_prompt}

\end{figure*}

\subsection{Longitudinal ICU Surveillance Agent}
\begin{figure*}[h]
    \begin{AIbox}{Longitudinal Surveillance System Prompt}

\begin{quote}
\ttfamily
\tiny
\vspace{2mm}

You are a general ICU rolling surveillance agent operating in rolling checkpoint mode.
This is a rolling monitoring task, not a forecasting task.
At each checkpoint, visible data and function outputs only contain information available up to the current checkpoint.

We monitor the following disease families:
\begin{itemize}[nosep]
\item infection and sepsis
\item renal injury and urine-output failure, including CRRT when relevant
\item respiratory support escalation and hypoxemia
\item hemodynamic instability, vasoactive support, and shock
\item neurologic deterioration
\item metabolic failure, including lactate elevation and acidemia
\item coagulation abnormality
\end{itemize}

You are given the current checkpoint step index and time, a `rolling\_history' summarizing previous checkpoints, an embedded surveillance guideline digest, and a catalog of patient-state functions.

At each turn, return exactly one of:
\begin{enumerate}[nosep]
\item a \textbf{tool call} to one focused patient-state function, when the current state is not yet sufficiently supported by `rolling\_history' or current-checkpoint evidence;
\item a \textbf{final decision} JSON, when the current state is sufficiently supported.
\end{enumerate}
Do not return a final decision merely because no patient evidence has been retrieved. ``Continue monitoring'' is a clinical decision and must also be evidence-supported.

\textbf{Decision semantics:}
\begin{itemize}[nosep]
\item \texttt{suspected\_conditions} --- clinically meaningful concern that should keep monitoring focused on that condition family.
\item \texttt{alerts} --- higher-acuity or higher-confidence states that justify escalation now.
\item \texttt{global\_action} --- exactly one of \texttt{continue\_monitoring} or \texttt{escalate}. If \texttt{alerts} is non-empty, \texttt{global\_action} should be \texttt{escalate}.
\item \texttt{priority} --- exactly one of \texttt{low}, \texttt{medium}, \texttt{high}.
\end{itemize}

\textbf{Final decision JSON contract:}
\{"global\_action":"continue\_monitoring$|$escalate",

"suspected\_conditions":["..."],

"alerts":["..."],

"priority":"low$|$medium$|$high",

"rationale":"..."\}

\textbf{Embedded guideline digest (excerpt):}
\begin{itemize}[nosep]
\item \emph{Infection / Sepsis:} treat infection as an episode-level state once credible evidence appears; escalate to sepsis when infection is present and organ dysfunction (Sepsis-3 evidence or shock physiology) is strong.
\item \emph{Renal:} interpret AKI by KDIGO stage and remember the worst stage so far; oliguria is a rolling-window short-horizon signal; CRRT indicates active support, not a permanent state.
\item \emph{Respiratory:} HFNC/NIV is support-escalation; invasive ventilation is alert-level. PF ratio, oxygenation, lactate, pH expire if not refreshed.
\item \emph{Hemodynamic:} any vasoactive support is a strong instability signal; shock-like states depend on combined evidence (sepsis, support, metabolic).
\item \emph{Neurologic:} GCS $\leq$ 8 is a severe deterioration signal; GCS reflects recent observed status.
\item \emph{Metabolic:} lactate $\geq$ 2 = stress, $\geq$ 4 = alert; pH $<$ 7.30 = acidemia, $\leq$ 7.20 = severe.
\item \emph{Coagulation:} INR $\geq$ 1.5 = meaningful, $\geq$ 2.0 = alert-level coagulopathy.
\end{itemize}

\textbf{Available patient-state functions} (callable as \{"tool\_name":"call\_function",

"arguments":\{"function\_name":"...",

"arguments":\{"stay\_id":N\}\}\}):
\texttt{get\_suspicion\_of\_infection}, \texttt{compute\_sofa\_score}, \texttt{kdigo\_stages}, \texttt{get\_urine\_output\_rate}, \texttt{ventilation\_info}, \texttt{get\_vasoactive\_agent\_info}, \texttt{gcs}, \texttt{get\_blood\_gas\_info}, \texttt{get\_coagulation}, \texttt{get\_crrt\_info}.

A \texttt{run\_python} fallback is available for focused custom queries against the checkpoint-scoped DuckDB session when \texttt{call\_function} does not return enough information.

\end{quote}

\end{AIbox}
    \caption{System prompt for the longitudinal ICU surveillance agent. The agent is invoked at every 4-hour checkpoint and must commit to either a focused tool call or a final structured monitoring decision. The full label vocabulary and the per-family guideline digest are embedded so the model has consistent definitions across checkpoints.}
    \label{prompts:surveillance_system_prompt}
\end{figure*}

\begin{figure*}[h]
    \begin{AIbox}{Longitudinal Surveillance User Prompt}

\begin{quote}
\ttfamily
\tiny
\vspace{2mm}

The user message is a single JSON object describing the current checkpoint and the agent's prior actions for this stay:

\vspace{1mm}
\{\\
\hspace*{2mm}"step\_input": \{\\
\hspace*{4mm}"trajectory\_id": "\{trajectory\_id\}",\\
\hspace*{4mm}"stay\_id": \{stay\_id\},\\
\hspace*{4mm}"step\_index": \{step\_index\},\\
\hspace*{4mm}"t\_hour": \{t\_hour\},\\
\hspace*{4mm}"task\_name": "general\_icu\_surveillance"\\
\hspace*{2mm}\},\\
\hspace*{2mm}"tool\_backend": "\{tool\_backend\}",\\
\hspace*{2mm}"available\_tools": [\{available\_tools\}],\\
\hspace*{2mm}"already\_called\_tools": [\{tools\_called\_this\_checkpoint\}],\\
\hspace*{2mm}"tool\_outputs\_in\_order": [\{tool\_outputs\_so\_far\}],\\
\hspace*{2mm}"repeated\_calls": [\{repeated\_call\_warnings\}],\\
\hspace*{2mm}"rolling\_history": \{\\
\hspace*{4mm}"0":  "...short summary at t=0...",\\
\hspace*{4mm}"4":  "...short summary at t=4...",\\
\hspace*{4mm}...\\
\hspace*{4mm}"\{t\_hour - 4\}": "...summary at the previous checkpoint..."\\
\hspace*{2mm}\}\\
\}
\vspace{2mm}

\textbf{Field semantics:}
\begin{itemize}[nosep]
\item \texttt{t\_hour} --- the current checkpoint, in hours since ICU admission. The agent is run for $t \in \{0, 4, 8, \dots, 48\}$.
\item \texttt{rolling\_history} --- a compact natural-language summary the agent itself produced at every \emph{prior} checkpoint of this stay. \texttt{null} entries mean a family was not yet assessed and must not be treated as negative evidence.
\item \texttt{already\_called\_tools} / \texttt{tool\_outputs\_in\_order} --- the tool calls and tool outputs the agent has already issued \emph{within the current checkpoint}, so it can chain evidence without re-querying.
\item \texttt{repeated\_calls} --- a nudge listing tool calls already issued at this checkpoint; the agent should not repeat them and should fall back to \texttt{run\_python} if more evidence is needed.
\end{itemize}

The model must respond with exactly one tool call or one final decision JSON, conforming to the system prompt's contract. After every checkpoint the model also writes a one-sentence \texttt{checkpoint\_summary} that becomes the next step's \texttt{rolling\_history} entry.

\end{quote}

\end{AIbox}
    \caption{User prompt for the longitudinal ICU surveillance agent at each 4-hour checkpoint. The payload is rendered as JSON; \texttt{rolling\_history} accumulates the agent's own per-checkpoint summaries from earlier in the stay, so the model has direct access to its prior decisions when reasoning about the current state.}
    \label{prompts:surveillance_user_prompt}
\end{figure*}

\clearpage


\section{Training Details}
\label{app:train_details}

\begin{table}[h]
\centering
\small
\begin{tabular}{ll}
\toprule
\textbf{Category} & \textbf{Hyperparameter} \\
\midrule

\multicolumn{2}{c}{\textit{Model}} \\
Max prompt length & 1024 \\
Max response length & 16384 \\
Gradient checkpointing & Enabled \\
Model dtype & bfloat16 \\

\midrule
\multicolumn{2}{c}{\textit{Training}} \\
Algorithm & GRPO \\
Total epochs & 1 \\
Train batch size & 4 \\
PPO mini-batch size & 4 \\
Actor learning rate & $1 \times 10^{-5}$ \\
Clip ratio (low) & 0.20 \\
Clip ratio (high) & 0.28 \\
Clip ratio constant & 10.0 \\
Use KL in reward & False \\
KL coefficient & 0.0 \\
Use KL loss & False \\
KL loss coefficient & 0.0 \\
Dynamic batch size & Enabled \\

\midrule
\multicolumn{2}{c}{\textit{Rollout / Inference}} \\
Backend & vLLM (async) \\
Tensor parallel size & 2 \\
Responses per prompt (train) & 4 \\
Responses per prompt (val) & 1 \\
Temperature (val) & 1.0 \\
Top-p (val) & 0.6 \\
Max user turns & 5 \\
Max assistant turns & 5 \\
Multi-turn format & Hermes \\

\bottomrule
\end{tabular}
\caption{Training and inference hyperparameters for the MIMIC-IV agent using GRPO.}
\label{tab:hyperparams}
\end{table}

\begin{figure}[!t]
    \centering

    \begin{subfigure}[t]{0.48\linewidth}
        \centering
        \includegraphics[width=\linewidth]{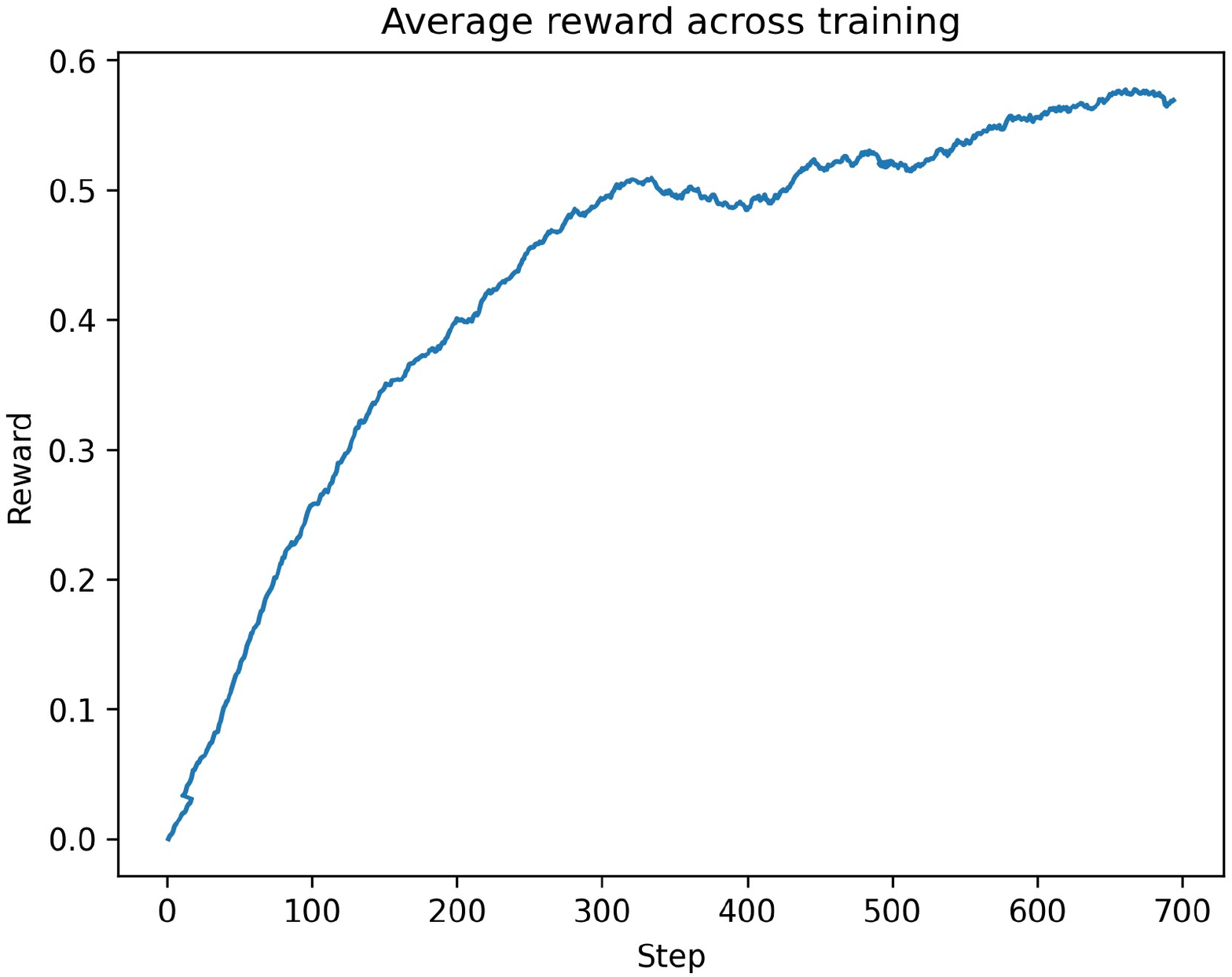}
        \caption{Average rewards across levels during GRPO training. We see a consistent trend of improved performance with more training samples.}
        \label{fig:grpo_progress}
    \end{subfigure}
    \hfill
    \begin{subfigure}[t]{0.48\linewidth}
        \centering
        \includegraphics[width=\linewidth]{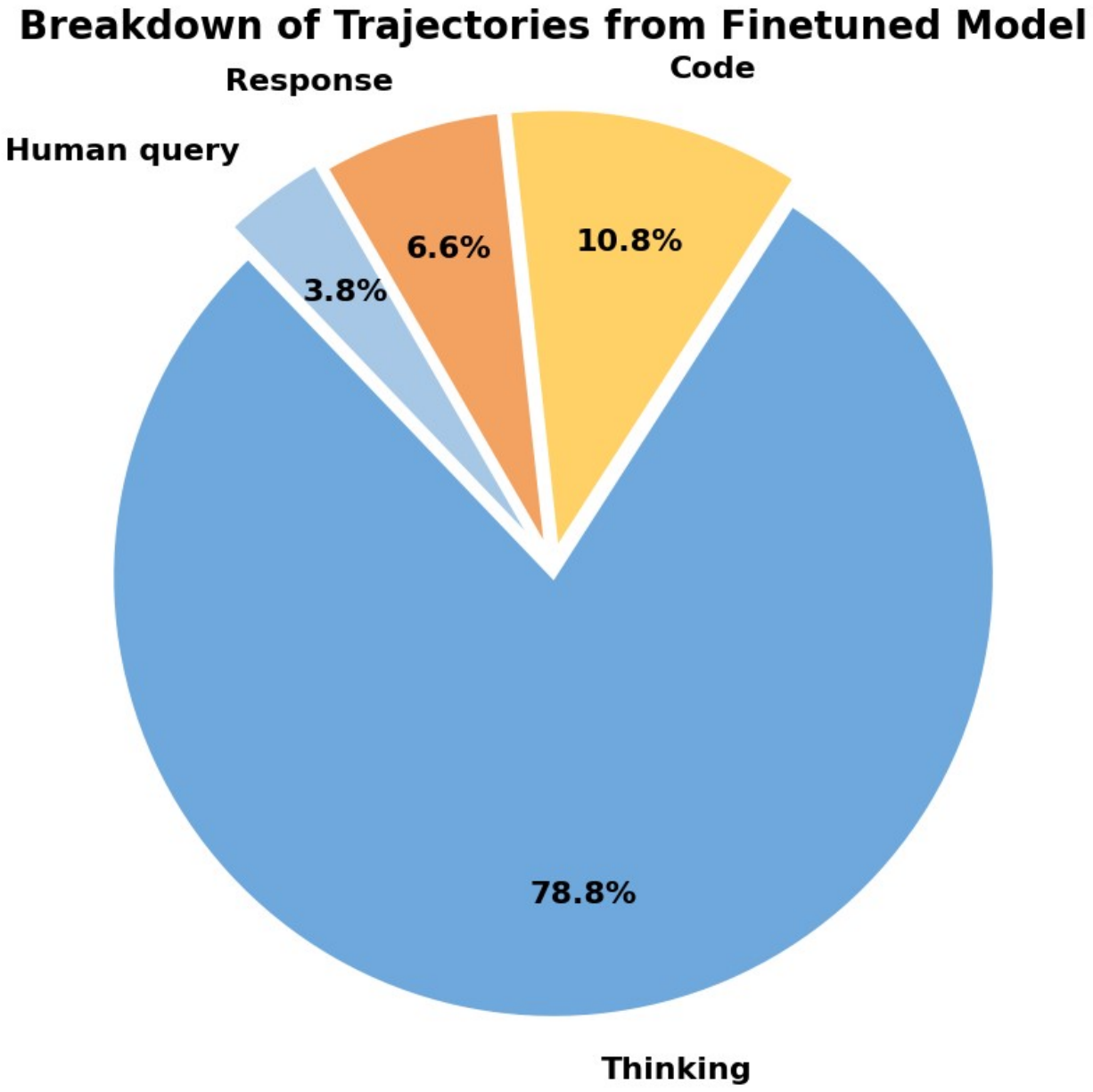}
        \caption{Distribution of responses from finetuned model.}
        \label{fig:finetuned_piechart}
    \end{subfigure}

    \caption{Findings from GRPO finetuning.}
    \label{fig:grpo_findings}

\end{figure}

Figure \ref{fig:grpo_progress} shows the training progress using GRPO with our dataset. We believe more long-term training using our dataset would lead to strong gains in model performance. Figure \ref{fig:finetuned_piechart} highlights a majority of token budget is spent on reasoning, followed by generating code. 

\clearpage


\section{Compute Resources}
\label{app:compute}

All inference experiments were run on a \texttt{8xB200 Node}. Inference was served with \texttt{vLLM}. Approximate runtimes per experimental condition are as follows:

\begin{itemize}[nosep]
    \item \textbf{Autoformalization (per concept):} 20 GPU-hours on average for a \texttt{Qwen3.5-27B} model run with efficient parallelization via topological ordering of concepts.
    \item \textbf{GRPO Finetuning:} 144 GPU-hours total on a subset of the training split (50 questions per concept) for a total of 3150 questions. \texttt{4xH200} GPUs were used for training.
    \item \textbf{zeroshot inference evaluation (56{,}700 queries, one model):} 114 GPU-hours when using \texttt{Qwen3.5-27B} and parallel processing of 4 concepts at a time. 
\end{itemize}

Total compute across all reported experiments (all models, all baselines) is estimated at $\sim$ 1000-2000 GPU-hours. No substantial additional compute was used for preliminary or failed experiments beyond the reported runs.

\section{Sample Inference Trajectory}
\label{app:sample trajectory}

We present a representative inference trajectory drawn from the MIMIC-IV Demo 
dataset (publicly available, no credentialing required).\footnote{MIMIC-IV Demo 
is available at \url{https://physionet.org/content/mimic-iv-demo/} under the 
PhysioNet Open Data Commons License.} The trajectory uses 
\texttt{Qwen/Qwen3.5-27B-Instruct} as the inference model, consistent with the 
backbone used in the main experiments (Section~5). No pre-computed derived tables 
are available and no guideline text is pre-injected into the prompt --- the agent 
must discover, retrieve, and apply clinical knowledge entirely through the guideline 
tools.
\begin{table}[h]
\centering
\small
\caption{Summary of the sample inference trajectory.}
\label{tab:trajectory-summary}
\begin{tabular}{ll}
\toprule
\textbf{Field} & \textbf{Value} \\
\midrule
Concept          & SOFA --- CNS component (GCS-based) \\
Question         & \textit{``What is this patient's CNS (GCS-based) SOFA} \\
                 & \textit{component score during their first 24 hours in the ICU?''} \\
Patient          & subject\_id=10007928, hadm\_id=20338077, stay\_id=35128235 \\
Ground truth     & \textbf{0} (GCS = 15, full consciousness) \\
Predicted        & \textbf{0} $\checkmark$ \\
LLM turns        & 7 \\
SQL queries      & 2 \\
Derived tables   & None \\
Guideline pre-injected & No \\
\bottomrule
\end{tabular}
\end{table}

\definecolor{rowshade}{RGB}{235,237,245}
\definecolor{headercolor}{RGB}{45,45,60}
\newcommand{\agentrow}[1]{%
  \begin{tcolorbox}[
    colback=rowshade,
    colframe=rowshade,
    boxrule=0pt,
    arc=0pt,
    left=4pt, right=4pt,
    top=3pt, bottom=3pt,
    before skip=4pt,
    after skip=0pt
  ]
  \small #1
  \end{tcolorbox}%
}

\newcommand{\inforow}[1]{%
  \vspace{2pt}
  {\small #1}
  \vspace{4pt}
}
\begin{figure}[p]   
\begin{tcolorbox}[
  enhanced,
  title={\textbf{Trajectory Example: CNS SOFA Component Score 
         (MIMIC-IV Demo)}},
  colbacktitle=headercolor,
  coltitle=white,
  fonttitle=\small\bfseries,
  colback=white,
  colframe=headercolor,
  boxrule=0.8pt,
  left=6pt, right=6pt,
  top=4pt, bottom=4pt,
  arc=3pt
]

\small
\textbf{Patient Query:}\quad
\textit{``What is this patient's CNS (GCS-based) SOFA component 
score during their first 24 hours in the ICU?''}
\hfill
\textbf{Label\,/\,Predicted:} 0\,/\,0\;$\checkmark$

\tcbline

\centering\small\textbf{Agent Trajectory}
\par\vspace{2pt}\raggedright

\agentrow{%
  \textbf{[Search]:}\quad
  \texttt{search\_guidelines(}\\
  \quad\texttt{"What is this patient's CNS (GCS-based)}\\
  \quad\texttt{SOFA score during their first 24 hours in the ICU?")}%
}
\inforow{%
  \textbf{[Information]:}\;
  \texttt{[`first\_day\_sofa\_guideline',\;
  `saps-ii\_guideline',\;
  `coma\_altered\_consciousness\_icu\_guideline',\;\ldots]}\\[2pt]
  Semantic retrieval ranks \texttt{first\_day\_sofa\_guideline} 
  first over 351 section-level embeddings. Full clinical 
  question used as query --- not a short keyword.%
}

\agentrow{%
  \textbf{[Read]:}\quad
  \texttt{get\_guideline("first\_day\_sofa\_guideline")}%
}
\inforow{%
  \textbf{[Information]:}\; 500-char preview + section menu 
  (not the full $\sim$6,000-token document):\\[3pt]
  \texttt{\small
  \quad--\;Definition \quad
  --\;Diagnostic Criteria \quad
  --\;\textbf{Scoring System}\\
  \quad--\;\textbf{Time Windows} \quad
  --\;Severity Staging \quad
  --\;Operational Notes}%
}

\agentrow{%
  \textbf{[Read Section]:}\quad
  \texttt{get\_guideline\_section(}%
  \texttt{"first\_day\_sofa",\;"Scoring System")}%
}
\inforow{%
  \textbf{[Information]:}\; Exact GCS$\to$score mapping 
  retrieved ($\sim$500 tokens):\\[4pt]
  \centering
  \footnotesize
  \begin{tabular}{lrrrrr}
  \toprule
  \textbf{GCS total}    & 15 & 13--14 & 10--12 & 6--9 & $<$6 \\
  \textbf{SOFA CNS score} & 0 & 1 & 2 & 3 & 4 \\
  \bottomrule
  \end{tabular}
  \par\raggedright%
}

\agentrow{%
  \textbf{[Read Section]:}\quad
  \texttt{get\_guideline\_section(}%
  \texttt{"first\_day\_sofa",\;"Time Windows")}%
}
\inforow{%
  \textbf{[Information]:}\;
  \textit{``Use the \textbf{lowest} GCS documented within the 
  24-hour window (worst-value selection).''}%
}

\agentrow{%
  \textbf{[Query\;DB]:}\quad
  \texttt{SELECT intime FROM icustays}\\
  \quad\texttt{WHERE stay\_id = '35128235'}
  \quad$\Rightarrow$\quad
  \texttt{2129-04-06\;00:25:00}%
}

\agentrow{%
  \textbf{[Query\;DB]:}\quad
  \texttt{SELECT itemid, valuenum FROM chartevents}\\
  \quad\texttt{WHERE stay\_id = '35128235'}\\
  \quad\texttt{AND itemid IN (220739, 223900, 223901)}\\
  \quad\texttt{AND charttime WITHIN first 24h}%
}
\inforow{%
  \textbf{[Information]:}\;
  Eye\,=\,4,\; Verbal\,=\,5,\; Motor\,=\,6 
  at all 4 timepoints.\;
  Total GCS\,=\,15 (consistent throughout window).%
}

\tcbline

\small
\textbf{[Think]:}\;
From scoring table (Turn~3): GCS\,=\,15\;$\to$\;SOFA 
CNS\,=\,0. From time window rule (Turn~4): minimum GCS 
used --- all readings are 15, so minimum\,=\,15.

\smallskip
\textbf{[Answer]:}\;
SOFA CNS score\,=\,\textbf{0}.\quad
\textit{The agent retrieved only the two sections it needed 
(Scoring System and Time Windows) rather than loading the 
full guideline document.}

\end{tcolorbox}
\caption{Sample inference trajectory on the MIMIC-IV Demo 
dataset. No guideline text is pre-injected; no derived 
tables are used. Model: \texttt{Qwen/Qwen3.5-27B-Instruct}.}
\label{fig:trajectory}
\end{figure}

\end{document}